\documentclass[lettersize,journal]{IEEEtran}
\usepackage{amsmath,amsfonts}
\usepackage{algorithmic}
\usepackage{array}
\usepackage[caption=false,font=normalsize,labelfont=sf,textfont=sf]{subfig}
\usepackage{textcomp}
\usepackage[colorlinks,citecolor=blue,bookmarks=false,hypertexnames=true]{hyperref}
\usepackage{stfloats}
\usepackage{url}
\usepackage{verbatim}
\usepackage{graphicx}
\usepackage{booktabs}
\usepackage{multirow}

\usepackage{balance}
\usepackage{algorithmic}
\usepackage[ruled]{algorithm2e}
\begin{document}
\title{VideoPure: Diffusion-based Adversarial Purification for Video Recognition}
\author{Kaixun~Jiang, Zhaoyu Chen, Jiyuan Fu, Lingyi Hong, Jinglun Li, and Wenqiang~Zhang~\IEEEmembership{Member,~IEEE}
\thanks{This work was supported by National Natural Science Foundation of China (No.62072112), Scientific and Technological innovation action plan of  Shanghai Science and Technology Committee (No.22511102202).}
\thanks{Kaixun Jiang, Zhaoyu Chen, Jinglun Li and Wenqiang Zhang are with Shanghai Engineering Research Center of AI Robotics, Academy for Engineering \& Technology, Fudan University, Shanghai, China, and also with Engineering Research Center of AI \& Robotics, Ministry of Education, Academy for Engineering \& Technology, Fudan University, Shanghai, China. Jiyuan Fu, Lingyi Hong, Wenqiang Zhang are with Shanghai Key Lab of Intelligent Information Processing, School of Computer Science, Fudan University, Shanghai, China. The emails of these authors are kxjiang22@m.fudan.edu.cn, zhaoyuchen20@fudan.edu.cn, \{fujy23, lyhong22, jinglunli21\}@m.fudan.edu.cn, wqzhang@fudan.edu.cn.

 (Corresponding author: Zhaoyu Chen and Wenqiang Zhang.)
}

}

\markboth{}%
{}
\maketitle

\begin{abstract}
Recent work indicates that video recognition models are vulnerable to adversarial examples, posing a serious security risk to downstream applications. However, current research has primarily focused on adversarial attacks, with limited work exploring defense mechanisms. Furthermore, due to the spatial-temporal complexity of videos, existing video defense methods face issues of high cost, overfitting, and limited defense performance. Recently, diffusion-based adversarial purification methods have achieved robust defense performance in the image domain. However, due to the additional temporal dimension in videos, directly applying these diffusion-based adversarial purification methods to the video domain suffers performance and efficiency degradation. To achieve an efficient and effective video adversarial defense method, we propose the first diffusion-based video purification framework to improve video recognition models' adversarial robustness: VideoPure. Given an adversarial example, we first employ temporal DDIM inversion to transform the input distribution into a temporally consistent and trajectory-defined distribution, covering adversarial noise while preserving more video structure. Then, during DDIM denoising, we leverage intermediate results at each denoising step and conduct guided spatial-temporal optimization, removing adversarial noise while maintaining temporal consistency. Finally, we input the list of optimized intermediate results into the video recognition model for multi-step voting to obtain the predicted class. We investigate the defense performance of our method against state-of-the-art black-box, gray-box, and adaptive attacks on benchmark datasets and models. Compared with other adversarial purification methods, our method overall demonstrates better defense performance against different attacks. Moreover, our method can be applied as a flexible defense plugin for video recognition models. {Our code is available at https://github.com/deep-kaixun/VideoPure}
\end{abstract}


\section{Introduction}
\IEEEPARstart{D}{eep} neural networks (DNNs) have exhibited remarkable performance in various computer vision tasks~\cite{he2016deep,chen2022towards}. However, recent studies~\cite{pgd,tcsvt1,tcsvt2,tcsvt3,tcsvt4,tcsvt5,tcsvt6} have exposed a fundamental vulnerability inherent in DNNs' capabilities, especially when facing adversarial examples. Adversarial examples introduce malicious and imperceptible perturbations to clean data, leading DNNs to make erroneous predictions with high confidence. Moreover, recent research has demonstrated the effectiveness of adversarial examples against video recognition models~\cite{SAP}. Actually, compared with images, videos as a higher-dimensional data modality, expose a larger attack surface, making video recognition models more susceptible to adversarial attacks. This vulnerability poses serious security risks to related downstream tasks such as autonomous driving~\cite{drive}, video surveillance~\cite{video_sur}, etc. Therefore, it becomes urgent to study video adversarial defense methods to enhance the security of video recognition models.

Unfortunately, current adversarial defense methods primarily focus on the image domain, while the video domain has received less attention. There are two main categories of adversarial defense methods. The first is adversarial training~\cite{pgd,at1,at2,at3,at4}, where adversarial examples are used as a form of data augmentation to retrain the model so that it learns more robust distribution. However, adversarial training has several drawbacks: 1) Impact and inflexibility. It necessitates modifying the model's intrinsic parameters, impacting the model's accuracy with clean data and compromising its flexibility. 2) Overfitting. Adversarial training demonstrates poor defense performance against unseen adversarial examples (not including training data). 3) High cost. Adversarial training incurs higher training costs compared to standard training, as it requires continuous generation of adversarial examples. This issue is even more pronounced for video recognition models due to their more complex spatial-temporal architectures. Consequently, research on adversarial training for video recognition models remains limited~\cite{bn_at,at_video}.

The second is adversarial purification~\cite{defense-gan,song2018pixeldefend,defend_video_pattern, jpeg,jpeg_video,tf,diffpure}, which treats adversarial perturbations as a form of noise and employs a pre-processing step to remove such noise before model inference. Unlike adversarial training, which requires modifying the model, adversarial purification only involves pre-processing the input during testing. Thus, minimal changes to the AI system can gain significant improvement in adversarial robustness. In the video domain, \cite{jpeg_video,defend_video_pattern,tf} have demonstrated the effectiveness of adversarial purification methods for defending video recognition models. However, when facing powerful adaptive attacks~\cite{bpda,EOT,autoattack}, these methods' defense performance decreases dramatically. Recently, diffusion-based adversarial purification methods~\cite{diffpure,blau2022threat,wu2022guided,wang2022guided,zhang2024enhancing} have exhibited robust defense performance in the image domain. These methods transform the adversarial image distribution into a simple Gaussian distribution through a diffusion process, and then remove the adversarial noise while retaining the clean image distribution through denoising process. As a result, these methods showcase strong defense performance on image classifiers. Hence, a question arises: \textit{Is it possible to transpose the diffusion-based purification defense onto video recognition models?}

A straightforward approach is to treat the video as a sequence of images, applying an image diffusion model to denoise each frame individually. The resulting denoised video is then fed into video recognition models. However, this method disregards the temporal correlation between frames, leading to sub-optimal defense performance (as shown in Table~\ref{tab:diffusion models}). Additionally, the iterative frame-by-frame processing, combined with DDPM sampling~\cite{ddpm} which has been commonly utilized in previous adversarial purification methods, significantly influences the efficiency.

To tackle the challenges mentioned above and develop an effective and efficient method for defending against adversarial attacks, we propose VideoPure, the first diffusion-based adversarial purification framework designed specifically for video recognition models. Our framework comprises three modules: temporal DDIM inversion, spatial-temporal optimization, and multi-step voting. Firstly, we improve efficiency by utilizing video diffusion models~\cite{wang2023modelscope} with efficient DDIM sampling~\cite{ddim} as the baseline, replacing traditional image diffusion models and DDPM sampling. Next, since recovering video structures is more challenging than images, we introduce temporal DDIM inversion for the diffusion process. This method propagates noise predicted from the initial frame temporally, replacing stochastic diffusion~\cite{ddpm}, thereby transforming the initial distribution into a temporally consistent and trajectory-defined distribution. Then, for the denoising process, we utilize intermediate results generated at each denoising timestep and perform spatial-temporal optimization on each intermediate result, preserving the video's original structure in the temporal domain while removing adversarial noise in the spatial domain. Finally, we input the list of optimized intermediate results into the video recognition model for multi-step voting to obtain the predicted class, thereby trapping the attack optimization into a local optimum for certain purification results of the voting list. Our method not only inherits the advantages of adversarial purification methods, operating independently of the video recognition model and allowing flexible application in downstream tasks, but it also achieves robust defense performance against multiple types of attacks. Our contributions can be summarized as follows:

\begin{itemize}
    \item To the best of our study, we are the first to introduce diffusion-based adversarial purification in the video domain and propose an effective and efficient framework VideoPure. To enhance defense efficiency, we adopt the video diffusion model and DDIM as the baseline, replacing traditional image diffusion models and DDPM.
    \item To improve defense performance, we propose temporal DDIM inversion for diffusion and spatial-temporal optimization for denoising, preserving the video's original structure while removing adversarial noise. Then, we predicted class through multi-step voting to trap the attack optimization in a local optimum.
    \item We verify the defense performance of our method against state-of-the-art black-box, gray-box, and adaptive attacks on benchmark datasets and models. Compared with other adversarial purification methods, our method demonstrates better defense performance.
    \item Our method requires no training or prior information of protected models. Therefore, it can be applied as a flexible defense plugin for video recognition models.
\end{itemize}

The remainder of the paper is organized as follows: Section~\ref{cap:relatedwork} briefly reviews the related work to adversarial attacks and adversarial defenses. Section~\ref{cap:method} first introduces the definition of diffusion-based video adversarial defense, and then details the proposed VideoPure framework. Section~\ref{cap:exp} shows the experimental results to demonstrate the effectiveness of our method, including performance comparison, ablation studies, visualizations, diagnostic experiments, effectiveness analysis, etc. Section~\ref{cap:con} concludes our work. 

\section{Related Work \label{cap:relatedwork}}
\subsection{Adversarial Attacks}
Goodfellow et al.~\cite{FGSM} first discover that introducing imperceptible perturbations to the original inputs can lead deep learning models to make incorrect predictions, such inputs are termed adversarial examples. Depending on the extent of information available to the attacker regarding the attacked model and defense method, adversarial attacks can be categorized as black-box attacks, gray-box attacks, and adaptive attacks.

\subsubsection{\textbf{Black-box Attacks}} Attackers can not access information about the attacked model or defense mechanisms. They generally need to employ surrogate models for designing adversarial examples to attack the target model (attacked model). On video recognition, TT~\cite{tt} utilizes video recognition models as surrogate models, aggregating gradients from different video input patterns to achieve effective attacks. GIE~\cite{gie}, I2V~\cite{I2V} employ image classification models as surrogate models, also achieving effective black-box attacks on video recognition models. Additionally, some video attack methods~\cite{V-BAD,SparkedPrior,jiang2023efficient,jiang2023towards,tcsvt3,tcsvt4} explore a soft black-box setting, wherein the attacker can access the model's output and optimize the attack based on it. However, as security awareness increases, it is becoming increasingly difficult for attackers to access the model's output multiple times~\cite{jiang2023efficient}. Hence, in this study, we primarily focus on the primary black-box setting, i.e., no information can be accessed from attacked models, selecting TT as the black-box attack to evaluate the defense method's performance. Black-box attacks are the most difficult attack setting for attackers, leading to the generation of adversarial examples that are generally lower in performance than gray-box attacks and adaptive attacks. Therefore it is often not possible to fully assess the performance of defense methods.

\subsubsection{\textbf{Gray-box Attacks}} Attackers can access all information about the attacked model but lack information about defense mechanisms. In the absence of adversarial defense considerations, gray-box attacks can be regarded as white-box attacks. Among them, gradient-based optimization methods, including FGSM~\cite{FGSM}, PGD~\cite{pgd}, and I-FGSM~\cite{I-FGSM}, have achieved strong attack performance by computing the gradients of the model with respect to the input to design adversarial examples. Taking PGD as an example, which is often used to evaluate the performance of adversarial defense, the generation process of adversarial examples can be represented as~\cite{pgd}:
\begin{equation}
    x_{adv}^{(i+1)}=\Pi_{x+\epsilon_{adv}}\left(x_{adv}^{(i)}+ a \operatorname{sgn}\left(\nabla_{x_{adv}^{(i)}} L(f_\theta(x_{adv}^{(i)}), y)\right)\right),
\label{eq:eq1}
\end{equation}
where $x_{adv}^{(i)}$ denotes the adversarial example generated in the $i^{th}$ iteration, $y$ denotes the label, $\Pi_{x+\epsilon_{adv}}$ ensures that $x_{adv}$ remains within the $\epsilon$-ball centered at the original input $x$, $a$ denotes the step size, $f_\theta(\cdot)$ denotes the attacked model, sgn$(\cdot)$ denotes the sign function and $\nabla_{x_{adv}} L(f_\theta(x_{adv}, y)$ denotes the gradient of loss with respect to $x_{adv}$. 

On video recognition, \cite{SAP} crafts adversarial examples for video recognition models through inter-frame perturbations and optimization with the Adam optimizer~\cite{kingma2014adam}. Leveraging the structural vulnerabilities inherent in video models,~\cite{video_attack_gray_box_1} effectively attacks by perturbing a single frame. Additionally,~\cite{filcking} introduces uniform RGB perturbations to each frame, mimicking variations in real-world lighting conditions. TT~\cite{tt} also can be seen as a gray-box attack when the surrogate model becomes the attacked model.

Consequently, gray-box attacks typically exhibit potent attack performance in the absence of defense methods, but their effectiveness significantly diminishes with the introduction of defense mechanisms. In this study, we employ PGD and TT as gray-box attacks to assess the performance of defense methods.

\subsubsection{\textbf{Adaptive Attacks}}
Attackers can access all information about the attacked model and defense methods. Therefore, attackers can treat the target model $f_\theta(\cdot)$ and the defense method $P(\cdot)$ as a new model $F_\theta(\cdot)=f_\theta(P(\cdot))$. Then, gray-box attacks such as PGD, can be utilized for adaptive attacks. For example, the adaptive attack form of PGD can be expressed as:
\begin{equation}
    x_{adv}^{(i+1)}=\Pi_{x+\epsilon_{adv}}\left(x_{adv}^{(i)}+ a \operatorname{sgn}\left(\nabla_{x_{adv}^{(i)}} L(F_\theta(x_{adv}^{(i)}), y)\right)\right),
\end{equation}
i.e., the gradient with respect to the attacked model $\nabla_{x_{adv}} L(f_\theta(x_{adv}, y)$  are transformed into the gradient with respect to the new model $\nabla_{x_{adv}} L(F_\theta(x_{adv}, y)$. However, some defense methods introduce non-differentiable operations (e.g. JPEG~\cite{jpeg}) or cause memory overflow during gradient computation (e.g. diffusion-based adversarial purification~\cite{diffpure}), preventing attackers from obtaining gradients.

To address this problem, Backward Pass Differentiable Approximation (BPDA)~\cite{bpda} replaces $P(\cdot)$ with an appropriate differentiable surrogate function $g(\cdot)$. For example, in adversarial purification, since the difference between the model's input and output mainly lies in the adversarial noise, the overall structure of the original input and the purified input remains largely consistent. Therefore, $g(\cdot)$ can be represented as $P(x)\approx g(x)=x$. The gradients $\nabla_{x_{adv}} L(F_\theta(x_{adv}), y)$ can then be transformed to $\nabla_{P(x_{adv})} L(f_\theta(P(x_{adv})), y)$.

Diffattack~\cite{diffattack} is a recently proposed adaptive attack method specifically designed for diffusion-based adversarial purification. To address memory issues, it employs a segment-wise gradient estimation method, reducing memory consumption to $O(1)$ cost. In order to mitigate the gradient vanishing or explosion problem caused by deep gradient chains, a residual-based reconstruction loss is proposed. Although Diffattack estimates more accurate gradients, the segmented gradient computation incurs additional time costs.

Many defense methods (e.g. diffusion-based) improve the defense performance by introducing randomness to bias the gradient estimated by the attacker when designing the attack. To deal with this situation, Expectation-Over-Transformation (EOT)~\cite{EOT} is proposed, which greatly improves the effectiveness of attacks against stochastic defenses by transforming the expected gradient, i.e., $\nabla_{x_{adv}}  \mathbb{E}[L(F_\theta(x_{adv}), y)] = \mathbb{E}[\nabla_{x_{adv}} L(F_\theta(x_{adv}), y)]$.

In summary, adaptive attacks represent the most challenging scenario for defense, thus, superior adaptive defense performance better demonstrates the efficacy of defense methods. In this study, we select PGD+BPDA, TT+BPDA, EOT+BPDA, and PGD+Diffattack to evaluate the performance of different defense methods against adaptive attacks from multiple perspectives.

\subsection{Adversarial Defense}
\subsubsection{\textbf{Adversarial Training}}
Adversarial training involves using adversarial examples as a form of data augmentation to retrain the model. This process helps the model learn a more robust decision boundary. Goodfellow et al.~\cite{fgsm_at} initially introduce adversarial examples during training by generating them using FGSM at each training step and injecting them into the training set. Subsequently, PGD gradually becomes the mainstream method for adversarial training~\cite{pgd}. \cite{bn_at} explores the impact of different forms of adversarial training on defense performance by employing multiple batch normalization layers.~\cite{at_video} explores the impact of different forms of adversarial training on the defense performance of video models. Adversarial training offers the advantage of significantly improving the model's defense capability against known adversarial examples. However, it remains vulnerable to unseen types of adversarial examples. Additionally, adversarial training on video-based models poses more challenges due to their high computational complexity~\cite{defend_video_pattern}.

\subsubsection{\textbf{Adversarial Purification}}
Adversarial purification involves pre-processing input data before feeding it into a model to eliminate adversarial noise. This pre-processing may include specific denoising techniques like JPEG~\cite{jpeg} or Wavelet Denoising (WD)~\cite{sr}, as well as generative denoising models such as Defense-GAN~\cite{defense-gan} and autoregressive generative models~\cite{song2018pixeldefend}. In the domain of videos,~\cite{jpeg_video} proposes an adaptive JPEG compression method based on optical flow. Additionally,~\cite{defend_video_pattern} introduces designed defense patterns into input videos to enhance adversarial defense performance. Temporal Shuffle~\cite{tf} defends against adversarial attacks by randomly shuffling the order of video frames, achieving notable defense performance. Recently, diffusion-based adversarial purification methods~\cite{diffpure,blau2022threat,wu2022guided,wang2022guided,zhang2024enhancing} have leveraged the potent generation and denoising capabilities of diffusion models, resulting in state-of-the-art robustness in image classification. Among them, Diffpure introduces random noise and deep gradient chains during the diffusion denoising process, rendering it resilient against advanced adversarial attacks. However, the high-dimensional attributes of video, including the additional temporal dimension, significantly increase both the difficulty and cost of defense. This results in performance and efficiency degradation when existing diffusion-based adversarial purification methods are applied to the video domain. We are the first to explore diffusion-based adversarial purification for video models and propose an effective and efficient diffusion-based framework for video adversarial purification.

\section{Method \label{cap:method}}
\subsection{Preliminary}
Given a clean video $x \in \mathbb{R}^{N \times H \times W \times C}$ and video recognition model $f_{\theta}(\cdot)$, where $N$, $H$, $W$, $C$ denote the number of frames, height, width, channel, respectively. We denote $\arg \max f_{\theta}(x)=y$ as the process of correct classification, where $y$ denotes the ground truth label of $x$.

We denote the adversarial video $x_{adv}$, on the one hand, $x_{adv}$ needs to deceive the video recognition model, i.e., $\arg \max f_{\theta}(x_{adv}) \neq y$, typically constrained by maximizing the loss. On the other hand, $x_{adv}$ also needs to be similar to $x$, typically constrained by $l_{p}$ distance, i.e., $\|x_{adv}-x\|_p$.

\textbf{Diffusion-based Adversarial Purification.} We represent the process of diffusion-based adversarial purification as $P(\cdot)$. Therefore, combing $P(\cdot)$, the defended video recognition model can be written as $f_{\theta}(P(\cdot))$. The optimization objective for defense can be expressed as:
\begin{equation}
    \underset{P(\cdot)}{\max} \sum_i \arg \max f_{\theta}(P(x^{i}))=y^{i}, \quad s.t. \quad  x^i \in \{x^i,x_{adv}^{i} \},
\end{equation}
where $(x^i,y^i)$ denotes $i^{th}$ test sample. $x_{adv}^{i}$ denotes the adversarial video generated by $x^{i}$. Therefore, the designed defense method $P(\cdot)$ should aim to minimally impact the recognition accuracy of clean data while enhancing the recognition accuracy of adversarial examples.

In this study, we employ a popular video-based latent diffusion model: ModelScope~\cite{wang2023modelscope} as the denoising model. Initially, ModelScope encodes the input video into a latent space, subsequently performs diffusion and denoising in the latent space, and finally decodes latent from latent space to pixel space. As a result, the process $P(\cdot)$ can be decomposed into four components: encoding  $\mathcal{E}(\cdot)$ , diffusion $q(\cdot)$ , denoising $p_\theta(\cdot)$, and decoding $\mathcal{D}(\cdot)$.

\begin{figure}[!t]
    \centering
    \scalebox{0.64}{
    \includegraphics{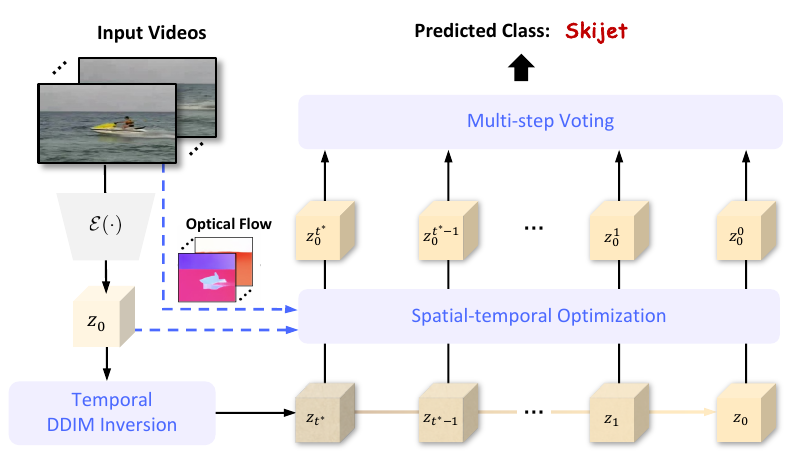}}
    \caption{Overview of our VideoPure framework. First, we encode video input to the latent space, getting $z_0$. And then we adopt temporal DDIM inversion to diffuse $z_0$ to $z_t$. For the denoising process, we fully utilize the intermediate results obtained from DDIM sampling, optimizing each $z_t^*$ in both spatial and temporal domains to derive the corresponding $z_0^t$. Finally, we input all optimized $z_0^t$ into the video recognition model after decoding, and aggregate the predictions through voting to obtain the final prediction category.}
    \label{fig:main}
\end{figure}
\textbf{Encoding \& Decoding.} The latent space in this paper is proposed by VQGAN~\cite{vqgan}, and the process of encoding and decoding can be represented as:
\begin{equation}
    z_{0} =\mathcal{E}(x), \quad x^* =\mathcal{D}({z}_{0}),
\end{equation}
where $z_{0} \in \mathbb{R}^{N \times \frac{H}{8} \times \frac{W}{8} \times 4}$ denotes the representation of the video in the latent space, $x^*$ denotes the purified $x$. Then, we remove the structure of $z_{0}$ by forward diffusion process $q(\cdot)$. 

\textbf{Diffusion.} Forward diffusion process can be seen as a Markov chain, commencing from the initial state $z_{0}$, gradually introducing small noise onto the $z_{0}$ to transform the complex input distribution into a simpler, well-defined distribution, such as the Gaussian distribution. This process can be formally defined as~\cite{ddpm}:
\begin{equation}
\begin{aligned}
q\left({z}_{1: t^*} \mid {z}_0\right)&:=\prod_{t=1}^{t^*} q\left({z}_t \mid {z}_{t-1}\right), \\ 
q\left({z}_t \mid {z}_{t-1}\right)&:=\mathcal{N}\left({z}_t ; \sqrt{1-\beta_t} {z}_{t-1}, \beta_t {I}\right),
\end{aligned}
\end{equation}
where $t^*$ denotes the total timesteps of diffusion, $\beta_t$ denotes forward process variance of $t^{th}$ timestep, which is generally a small constant that increases linearly~\cite{ddpm}. The above recursive equation can be simplified as follows~\cite{ddpm}:
\begin{equation}
\begin{aligned}
q({z}_{t^*} \mid {z}_0)=\mathcal{N}({z}_{t^*} ; \sqrt{\bar{\alpha}_{t^*}} {z}_0,(1-\bar{\alpha}_{t^*}) {I}),\\
    {z}_{t^*}=\sqrt{\bar{\alpha}_{t^*}} \cdot {z}_0+\sqrt{1-\bar{\alpha}_{t^*}} \cdot \epsilon, \quad \epsilon \sim \mathcal{N}({0},{I}),
\label{eq:diffusion}
\end{aligned}
\end{equation}
where $\alpha_{t^*}=1-\beta_{t^*}$ and $\bar{\alpha}_{t^*}= \prod_{t=1} \alpha_t$. Therefore, we only need  $z_{0}$ and the timestep $t^*$ to obtain the corresponding $z_{t^{*}}$. From the diffusion process of Eq.~\ref{eq:diffusion}, we can draw the following conclusions: 1) Since $\epsilon$ is sampled from a Gaussian distribution, the generation of $z_{t}$ exists randomness. 2) The larger $t^{*}$ is, the smaller $\sqrt{\bar{\alpha}_{t^*}}$ becomes, making $z_{t^{*}}$ closer to a Gaussian distribution. This implies that as adversarial noise is increasingly smoothed out, the disruption to the original structural information also increases.

\textbf{Denoising.} Reverse denoising process aims to remove the noise starting from $z_{t^*}$ to recover $z_{0}^*$. This process can be illustrated as follows:
\begin{equation}
p_\theta\left({z}_{t-1}^{*} \mid {z}_t^{*}\right)=\mathcal{N}\left({z}_{t-1}^{*} \mid \mu_\theta\left({z}_t^{*}, t\right),  \sigma_t^{2}I\right),
\end{equation}
where $z_{t}^{*}$ denotes the denoised latent at timestep $t$, amog them, $z_{t^*}^{*}=z_{t^{*}}$. $\mu_{\theta}(z_{t}^*,t)$ denotes the predicted mean, generated by diffusion model with $z_{t}^{*}$ and predicted noise $\epsilon_{\theta}(z_{t}^*,t) \in  \mathbb{R}^{N \times \frac{H}{8} \times \frac{W}{8} \times 4}$. $\sigma_t^{2}I$ denotes the $t^{th}$ timestep's variance. In general, $\sigma_t^{2}$ is often defined as a hyper-parameter~\cite{ddpm}. Taking DDPM sampling as an example, $\mu_{\theta}(z_t^*,t)=\frac{1}{\sqrt{\alpha_t}}\left({z}_t^*-\frac{1-\alpha_t}{\sqrt{1-\bar{\alpha}_t}} {\epsilon}_\theta\left({z}_t^*, t\right)\right)$. Since DDPM is predicated on a Markov chain process, it necessitates a step-by-step denoising operation, progressing from $z_{t^*}$ to reconstruct $z_0^*$. To enhance denoising efficiency, DDIM~\cite{ddim} is widely used as an efficient skip-step sampler. The formal representation of DDIM is illustrated as follows~\cite{ddim}:
\begin{equation}
\begin{aligned}
{z}_{t-1}^*=&\sqrt{\bar{\alpha}_{t-1}}\left(\frac{{z}_t^*-\sqrt{1-\bar{\alpha_t}} \epsilon_\theta\left({z}_t^*, t\right)}{\sqrt{\bar{\alpha_t}}}\right)\\
        &+\sqrt{1-\bar{\alpha}_{t-1}} \epsilon_\theta\left({z}_t^*, t\right),
\label{eq:ddim}
\end{aligned}
\end{equation}
DDIM is a non-Markov chain sampling technique that allows for skip-step sampling implementation. This means that it is not required to specify all $z_{1:t^*}^*$ step by step; instead, we can perform skip-step sampling within subsets of the process.

Previous research~\cite{diffpure} confirms that the distribution of adversarial examples and clean data will converge to a similar distribution after a certain timestep $t^*$ of diffusion. As the diffusion model is inherently trained on clean data, the denoised ${z}_{0}^*$ produced by DDPM or DDIM preserves the structure of $z_{0}$ while effectively removing the adversarial noise. Consequently, this process achieves a desirable defense effect. Due to the complexity of reconstructing video structures compared to images, we propose temporal DDIM inversion to transform the initial distribution of clean videos or adversarial videos into a temporally consistent and trajectory-defined simple distribution, helping reconstruct more original structural information during the denoising process. Additionally, we propose spatial-temporal optimization and multi-step voting to guide the denoising process. By optimizing multiple reconstructed video distributions in the spatial-temporal domain, confusing the attacker's optimization direction while robustly removing adversarial noise.

\subsection{VideoPure Framework}
\textbf{Overview.} Fig.~\ref{fig:main} provides an overview of our VideoPure framework, comprising three modules: temporal DDIM inversion, spatial-temporal optimization, and multi-step voting. Initially, we transform the video $x$ to $z_0$ from the pixel space to the latent space, utilizing temporal DDIM inversion for diffusion to obtain $z_{t^*}$. Subsequently, we employ DDIM sampling for denoising. During the denoising process, we fully exploit the intermediate results obtained at each denoising timestep, and optimize them by spatial-temporal optimization. Specifically, we disrupt the structural adversarial noise in the spatial domain while preserving the consistency of the video structure in the temporal domain. Finally, we aggregate all optimized intermediate results to achieve multi-step voting. Specifically, after decoding them back to the pixel space, we input them into the video recognition model. The prediction category with the highest number of votes is selected as the final prediction result.

\subsubsection{\textbf{Temporal DDIM Inversion}}

In the domain of real image editing~\cite{null,chen2023aca}, to preserve the structure of the original input, DDIM inversion~\cite{ddim} is widely adopted to transform the initial latent representation $z_{0}$ into a corresponding $z_{t^*}$ through a specific inversion scheme, rather than relying on the stochastic diffusion process described in Eq.~\ref{eq:diffusion}. Since the trajectory of DDIM inversion is deterministic, $z_{t^*}$ encoded via DDIM inversion can effectively reconstruct the details and structure of the initial latent representation $z_{0}$ after DDIM sampling. Therefore, to ensure maximal retention of recognition accuracy for clean data, we opt for DDIM inversion for latent encoding. Specifically, DDIM inversion first transforms Eq.~\ref{eq:ddim} as follows:
\begin{equation}
\begin{aligned}
{z}_{t}=&\sqrt{\bar{\alpha_{t}}}\left(\frac{{z}_{t-1}-\sqrt{1-\bar{\alpha}_{t-1}} \epsilon_\theta\left({z}_t, t\right)}{\sqrt{\bar{\alpha}_{t-1}}}\right)\\
        &+\sqrt{1-\bar{\alpha}_{t}} \epsilon_\theta\left({z}_t, t\right).
\label{eq:ddim_inversion1}
\end{aligned}
\end{equation}
Eq.~\ref{eq:ddim_inversion1} describes the process of generating $z_t$ from the encoding of $z_{t-1}$, which is the reverse process of denoising. However, every stage of encoding relies on the predicted noise of the next step $\epsilon_\theta(z_{t},t)$. As $z_{t}$ is not available, the predicted noise cannot be obtained. Hence, $\epsilon_\theta(z_{t-1},t)$ is adopted as a substitute, under the assumption that $z_{t-1}$ is approximately equal to $z_{t}$~\cite{edit_survey}. Through this alteration, the formal expression of the DDIM inversion is transformed as follows:
\begin{equation}
\begin{aligned}
{z}_{t}=&\sqrt{\bar{\alpha_{t}}}\left(\frac{{z}_{t-1}-\sqrt{1-\bar{\alpha}_{t-1}} \epsilon_\theta\left({z}_{t-1}, t\right)}{\sqrt{\bar{\alpha}_{t-1}}}\right)\\
        &+\sqrt{1-\bar{\alpha}_{t}} \epsilon_\theta\left({z}_{t-1}, t\right).
\label{eq:ddim_inversion}
\end{aligned}
\end{equation}

\begin{figure}[!t]
\centering
  \scalebox{0.5}{
  \includegraphics[width=0.9\textwidth]{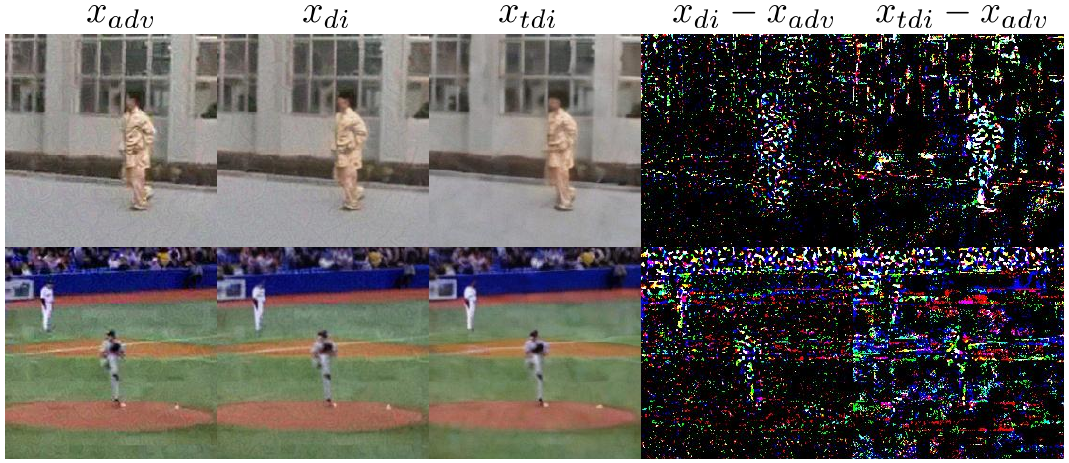}}
  \caption{Visualizations after different inversion methods. $x_{adv}, x_{di}, x_{tdi}$ denote adversarial examples generated by PGD, DDIM inversion, and temporal DDIM inversion. Upon comparison, the temporal DDIM inversion, evident in the second and third columns, eliminates more adversarial noise than the standard DDIM inversion. The fourth and fifth columns emphasize this further, showing more difference with $x_{adv}$. }
  \label{fig:ddim_inversion}
\end{figure}

However, while DDIM inversion guarantees the recognition accuracy of clean data, it restores too much structural information of $z_{0}$. As can be observed from Fig.~\ref{fig:ddim_inversion}, the sole use of DDIM inversion leads to the restoration of the detail of the adversarial noise, thus failing to achieve effective defense against adversarial examples. Based on this, we optimize DDIM inversion to suit the video adversarial purification. Specifically, motivated by the contributions of temporal noise initialization to video generation~\cite{khachatryan2023text2video}, we replace the noise added to each frame at each step with the first frame of predicted noise. That is, for the diffusion at timestep $t$, Eq.~\ref{eq:ddim_inversion} becomes:
\begin{equation}
\begin{aligned}
{z}_{t}=&\sqrt{\bar{\alpha_{t}}}\left(\frac{{z}_{t-1}-\sqrt{1-\bar{\alpha}_{t-1}} \epsilon_\theta^* \left({z}_{t-1}, t\right)}{\sqrt{\bar{\alpha}_{t-1}}}\right)\\
        &+\sqrt{1-\bar{\alpha}_{t}} \epsilon_\theta^* \left({z}_{t-1}, t\right),
\label{eq:temporal_ddim_inversion}
\end{aligned}
\end{equation}
where $\epsilon_\theta^*({z}_{t-1}, t)$ denotes the noise replicated from the predicted first frame and applied $N$ times to each subsequent frame. After this transformation, we can transform the initial distribution of clean videos or adversarial videos into a temporally consistent and trajectory-defined simple distribution. The advantage of temporal DDIM inversion lies in two main aspects. Firstly, compared to the stochastic diffusion in Eq.~\ref{eq:diffusion}, the diffusion trajectory of temporal DDIM inversion is more deterministic, which allows it to reconstruct more of the video's original structure after DDIM denoising. Secondly, unlike standard DDIM inversion, our method introduces a temporal difference between the diffusion trajectory and DDIM denoising trajectory. This temporal discrepancy enables the denoised video to retain substantial structural information while discarding finer details, such as adversarial noise. As illustrated in Fig.~\ref{fig:ddim_inversion}.

\subsubsection{\textbf{Spatial-temporal Optimization}}
\begin{figure}[!t]
\centering
  \scalebox{0.52}{
  \includegraphics[width=0.9\textwidth]{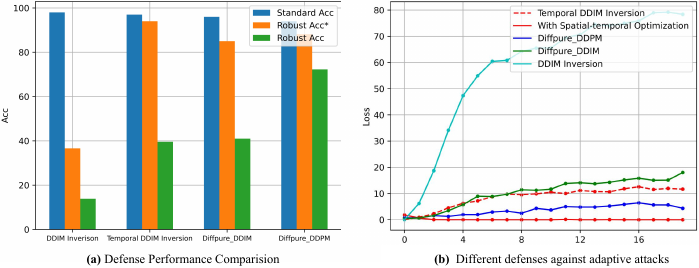}}
  \caption{(a) shows the performance comparison of different defense methods. (b) shows the loss of different defense methods in facing adaptive attacks, and the greater the rise of the loss, the worse the performance of the defense. PGD for evaluating Robust Acc* and PGD+BPDA for evaluating Robust Acc.}
  \label{fig:motivation}
\end{figure}

As shown in Fig.~\ref{fig:motivation}(a), temporal DDIM inversion greatly enhances the model's defense performance against gray-box attacks, i.e., high Robust Acc*. However, under the adaptive attack setting, where the attacker can access the details of defense methods, there is still poor defense performance compared with Diffpure. In the adaptive attack setting, attackers treat the defense model and video recognition model as a new model. For instance, in the case of PGD attacks, they can estimate the gradient of the combined model to implement an effective attack. Fig.~\ref{fig:motivation}(b) shows that temporal DDIM inversion under this attack setting indicates a linear ascent in loss with the number of iterations, failing to provide any defense effect. Therefore, considering the characteristics of videos and the optimization goal of the defense, we propose spatial-temporal optimization, aiming to optimize the denoised latent through the introduction of complex spatial-temporal noise patterns, making it difficult for attackers to find adversarial examples that consistently exceed the probability decision boundary.

In adversarial defense, there are two optimization objectives: firstly, when confronted with adversarial examples, the aim is to eliminate adversarial noise; secondly, when handling clean examples, it is essential to maintain structural similarity. Hence, we address these two optimization objectives separately in the spatial and temporal domains.

In the spatial domain, for each frame, the adversarial perturbation is attached to each pixel, thereby influencing the judgment of the classifier. As such, we aim to purify adversarial perturbations by distancing the denoising-generated ${z}_0^*$ from the original input-encoded $z_0$. Therefore, we employ the Mean Squared Error (MSE) loss as a constraint, which is explicitly represented as follows:
\begin{equation}
L_{spa}= -\| z_{0} - {z}_{0}^* \|_2.
\label{eq:l_spa}
\end{equation}
Through the optimization, the noise of each frame of the video is optimized to a variable degree, which increases the difficulty of the attack optimization. However, it's apparent that the spatial loss, not only disrupts adversarial noise, but also undermines the structure of original videos. Therefore, from the temporal domain, we ensure content consistency among adjacent frames, which results in maintaining the consistency of the temporal structure between the denoising video and the original video. Motivated by~\cite{yang2023motion}, we leverage temporal dynamics in videos to optimize the denoising latent. Specifically, we constrain the process by calculating the $L_1$ reconstruction loss between the flow-based warping of $z_0^*$ and $z_0^*$ itself, which can be represented as follows:
\begin{equation}
L_{temp}= \sum_{i=1}^{N-1} \|Warp\left({z}_0^*(i+1), O(i)\right)-{z}_{0}^*(i) \|_1,
\label{eq:l_temp}
\end{equation}
where $Warp({z}_0^*(i+1), O(i))$ denotes the warping frame of $i^{th}$ frame, i.e., backward warping, generated based on ${(i+1)}^{th}$ frame and the optical flow between $i^{th}$ and ${(i+1)}^{th}$ frame.

Combing $L_{temp}$, our loss can be described as $L=\lambda_1 L_{temp}+ \lambda_2 L_{spa}$. Finally, the process of spatial-temporal optimization can be represented as:
\begin{equation}
    {z}_0^*={z}_0^*-\alpha_s \nabla_{z^*_{0}}L,
\label{eq:optimization}
\end{equation}
{where $\alpha_s$ is the guidance scale. Hence, when confronted with adversarial attacks,} particularly adaptive attacks, each attack optimization iteration introduces complex spatial-temporal noise. This makes it challenging for the attacker to accurately estimate the gradient and consequently difficult to generate adversarial examples that consistently surpass the probability decision boundary, as shown in Fig.~\ref{fig:motivation} (b).

\begin{algorithm}[tb]
\caption{VideoPure}
\label{alg:algorithm}
\textbf{Input}: input video $x$, encoder $\mathcal{E}$, decoder $\mathcal{D}$, video recognition model $f_\theta$, denoising model $\epsilon_\theta$, diffusion timestep $t^*$\\ 
\textbf{Output}: predicted classification \\
\begin{algorithmic}[1]
\STATE  $vote\_list=\{\}$.
\STATE $z_{0}=\mathcal{E}(x).$\\
\STATE Generate $z_{t^*}$ using temporal DDIM inversion. \\
\FOR { $t$~in~$[t^*,1]$}
\STATE $z_0^t=\frac{{z}_t^*-\sqrt{1-\bar{\alpha}_t} \epsilon_\theta\left({z}_t^*, t\right)}{\sqrt{\bar{\alpha}_{t}}}$\\
\STATE ${z}_{t-1}^*=\sqrt{\bar{\alpha}_{t-1}}z_{0}^{t} +\sqrt{1-\bar{\alpha}_{t-1}} \epsilon_\theta\left({z}_t^*, t\right)$.\\
\STATE Calculate $L_{temp}$ and $L_{spa}$ with $z_{0}^{t}, z_0$.\\
\STATE  $z_t^{\prime}={z}_t^*-\alpha_s \sigma_t^2 \nabla_{z_{t}^*}L.$\\
\STATE  $z_0^t=\frac{{z}_t^{\prime}-\sqrt{1-\bar{\alpha}_t} \epsilon_\theta\left({z}_t^*, t\right)}{\sqrt{\bar{\alpha}_{t}}}$.\\
\STATE $vote\_list = vote\_list + \{z_0^t\}$.\\
\ENDFOR
\STATE Calculate $L_{temp}$ and $L_{spa}$ with $z_{0}^{*},z_{0}$.\\
\STATE  $z_0^{0}={z}_0^*-\alpha_s \sigma_t^2 \nabla_{z_{0}^*}L.$\\
\STATE  $vote\_list = vote\_list + \{z_0^0\}$.
\STATE $y_{pred}=[]$\\
\FOR {$z_0^t$~in~$vote\_list$}
\STATE $y_{pred} = y_{pred} + \arg \max f_\theta(\mathcal{D}(z_0^t))$.
\ENDFOR
\STATE \textbf{return} $\max y_{pred}$
\end{algorithmic}
\label{alg: algorithm}
\end{algorithm}

\subsubsection{\textbf{Multi-step Voting}}
Diffusion-based adversarial purification leverages the extremely deep gradient chain during inference to trap attackers in gradient explosion, vanishing, or memory explosion, thereby achieving effective defense~\cite{diffattack}. Inspired by this, we propose to design more complex gradient chains to further increase the difficulty of attack optimization. Specifically, we propose to broaden the gradient chain by utilizing the intermediate results of each denoising step and combine it with spatial-temporal optimization to further increase the difficulty of attack optimization, trapping attacks in a local optimum of a voting sample.

DDIM sampling in Eq.~\ref{eq:ddim} can be divided into two steps. The first step predicts the process of $z_{0}^t$ based on the current time step $z_t^*$ and $\epsilon_\theta(z_t^*,t)$, and the second step obtains $z_{t-1}^*$ by using Eq.~\ref{eq:diffusion}. Therefore, we can gain a series of $z_{0}^t$ during denosing process. Specifically, as $\epsilon_\theta(z_t^*,t)$ can be approximated as Gaussian noise, $\epsilon$ in Eq.~\ref{eq:diffusion} can be replaced by $\epsilon_\theta(z_t^*,t)$. Upon reshaping Eq.~\ref{eq:diffusion}, we can calculate the predicted $z_{0}^t$ based on timestep $t$:
\begin{equation}
   z_0^{t}= \frac{{z}_t^*-\sqrt{1-\bar{\alpha}_t} \epsilon_\theta\left({z}_t^*, t\right)}{\sqrt{\bar{\alpha}_{t}}}.
\label{eq:ddim_inverion_1}
\end{equation}
Therefore, in the denoising process, each timestep can yield its corresponding predicted $z_0^t$. To fully utilize this information, we carry out the above spatial-temporal optimization for each step $z_0^t$ of the denoising process. Specifically, $z_t$ first generates a corresponding $z_0^t$ based on Eq.~\ref{eq:ddim_inverion_1}. Next, we calculate the $L_{temp}$ and $L_{spa}$, and then optimize $z_{t}$ for the improvement of Eq.~\ref{eq:optimization}:
\begin{equation}
    {z}_t^{\prime}={z}_t^*-\alpha_s \sigma_t^2 \nabla_{z_{t}^*}L,
\end{equation}
{where $\sigma_t^2$ represents the variance at timestep $t$ in the DDIM schedule, which controls the degree of modification applied to the latent~\cite{yang2023motion}.} Then, we use ${z}_t^{\prime}$ to generate the optimized $z_{t}^{0}$. Finally, we obtain $(t^*+1)$ optimized $z_{0}^{t}$. After decoding all , they are inputted into the video recognition model to yield $(t^*+1)$ classification results. The category with the highest vote count is selected as the final classification result. Multi-step voting makes it possible for an attack optimization to get trapped in a local optimum by trapping a candidate solution ($z_0^t$) in the voting list, so multi-step voting can also improve the overall defense performance. Algorithm~\ref{alg: algorithm} describes the whole process of our VideoPure.

\section{Experiments \label{cap:exp}}
\subsection{Experimental Settings}

\textbf{Datasets.}
To evaluate the performance of our proposed framework, VideoPure, we employ two widely used video recognition datasets: UCF-101~\cite{UCF-101} and Kinetics-400~\cite{Kinetics-400}. The UCF-101 dataset comprises 13,320 video clips spanning 101 action categories. Following~\cite{tt,jiang2023efficient}, we adopt a random sampling approach, selecting one video from each category for evaluation. The Kinetics-400 dataset encompasses 400 action categories, with approximately 240,000 video clips for training and 20,000 for validation. Considering the computational cost of comparison methods (e.g., DiffPure), we randomly select 100 categories and one video from each category for evaluation. We ensure that all the selected videos can be correctly classified by all video recognition models.

\textbf{Models.}
To thoroughly assess the performance of our proposed framework, we conduct comprehensive evaluations on two widely used pretrained video recognition models: Non-local Network (NL)~\cite{NL} and SlowFast~\cite{slowfast}. For each video recognition model, we employ ResNet-50~\cite{he2016deep} as the backbone architecture. We standardize the input size for all models to $N \times H \times W \times C =32 \times 224 \times 224 \times 3$.

\textbf{Metrics.}
We evaluate the defense performance from three aspects: 1). Standard Acc(\%)~\cite{diffpure}: the recognition accuracy against clean data. 2). Robust Acc(\%)~\cite{diffpure}: the recognition accuracy against adaptive attacks.  3). Robust Acc*(\%): the recognition accuracy against gray-box attacks.

\textbf{Attack Methods.}
We select strong attack methods to evaluate our defense method. For gray-box attacks, we choose PGD~\cite{pgd} and TT~\cite{tt} which have strong attack performance for image and video models, respectively. For adaptive attacks, we select PGD, TT, Expectation Over Transformation (EOT)~\cite{EOT}, AutoAttack~\cite{autoattack} as attack methods, and we combine them with BPDA~\cite{bpda}, Diffattack~\cite{diffattack} to calculate defense methods' gradient, achieving effective adaptive attacks. We also select TT to verify the defense performance against black-box attacks.

\textbf{Defense Methods.}
We primarily select state-of-the-art adversarial purification methods in both the image and video domains for performance comparison. We initially opt for the classic JPEG~\cite{jpeg} and WD~\cite{sr} methods. Next, we select the efficient video defense method Temporal Shuffle~\cite{tf}. We then choose diffusion-based adversarial purification method Diffpure~\cite{diffpure}, comparing both its DDPM (Diffpure\_DDPM) and DDIM (Diffpure\_DDIM) versions. Additionally, we consider DDIM inversion~\cite{ddim} (without the inclusion of any modules we propose) as the baseline for our comparison. To ensure a fair comparison and demonstrate the effectiveness of our proposed method, we utilize the same video diffusion model ModelScope~\cite{wang2023modelscope} that our method adopts, as the denoising model for all compared approaches.

\textbf{Implement Details.}
For $l_\infty$ attacks, we set $\epsilon_{adv} = 4/255$, for $l_2$ attacks, we set $\epsilon_{adv}=2$. For most experiments we chose the most popular $l_\infty$ attacks to evaluate defense performance. We set attack step is $10$ for PGD and TT. We set $eot\_attack\_reps = 20$ for EOT. For AutoAttack, we follow the default $random$ setting. For the extra hyper-parameters in our method, we set timestep $t^* = 6$ for DDIM ($T = 50$), $t^* = 101$ for DDPM ($T = 1000)$, guidance scale $\alpha_s = -4$, and $\epsilon_t^2$ is followed~\cite{yang2023motion}, i.e., $\epsilon_t^2=log \frac{1-\bar{\alpha}_{t-1}}{1-\bar{\alpha}_t}\beta_t <0$. {To balance the range of $L$ ($\|l_{temp}\| \gg \|l_{spa}\|$),} we set $\lambda_1 = 5, \lambda_2 = 800$. Hyper-parameters tuning's detials are in Section~\ref{sec:hyper}. We adopt the RAFT model~\cite{teed2020raft} for optical flow estimation. Our method is implemented using Hugging Face\footnote{Hugging Face, \url{https://huggingface.co/ali-vilab/text-to-video-ms-1.7b}}.

\begin{table*}[!t]
\begin{center}
\caption{Comparison of defense performance of our method to other six defense methods under six attack methods. Average denotes the average value of the defense performance against the six attack methods.}
\begin{tabular}{@{}c|c|c|cccccccc@{}}
\toprule
 &  &  & \multicolumn{8}{c}{Attack Method} \\ \cmidrule(l){4-11} 
\multirow{-2}{*}{Dataset} & \multirow{-2}{*}{Model} & \multirow{-2}{*}{Defense Method} & \multicolumn{1}{c|}{None} & PGD & PGD+BPDA & TT & TT+BPDA & EOT+BPDA & PGD+Diffattack & Average \\ \midrule
 &  & None & \multicolumn{1}{c|}{100} & 8.9 & - & 11.8 & - & - & - & - \\
 &  & JPEG & \multicolumn{1}{c|}{99} & 8.9 & 9.9 & 13.9 & 16.8 & 9.9 & - & 11.8 \\
 &  & WD & \multicolumn{1}{c|}{100} & 7.9 & 9.9 & 12.9 & 12.9 & 9.9 & - & 10.7 \\
 &  & Temporal Shuffle & \multicolumn{1}{c|}{95} & 84.1 & 12.87 & 12.9 & 9.9 & 5.9 & - & 25.1 \\
 &  & Diffpure\_DDPM & \multicolumn{1}{c|}{94} & \underline{88.2} & \underline{72.3} & \textbf{72.3} & \underline{62.4} & \underline{39.6} & \underline{65.4} & \underline{66.7} \\
 &  & Diffpure\_DDIM & \multicolumn{1}{c|}{96} & 85.1 & 41.6 & 54.5 & 54.5 & {13.9} & {22.8} & 45.4 \\
 &  & DDIM Inversion & \multicolumn{1}{c|}{98} & 36.6 & 13.9 & 21.8 & 18.8 & {12.9} & {22.8} & 21.1 \\
 & \multirow{-8}{*}{NL} & Ours & \multicolumn{1}{c|}{96} & \textbf{94.1} & \textbf{100} & \underline{55.4} & \textbf{80.2} & \textbf{97.0} & \textbf{89.1} & \textbf{85.9} \\ \cmidrule(l){2-11} 
 &  & None & \multicolumn{1}{c|}{100} & 7.9 & - & 5.0 & - & - & - & - \\
 &  & JPEG & \multicolumn{1}{c|}{99} & 8.9 & 7.9 & 5.9 & 8.9 & 11.9 & - & 8.7 \\
 &  & WD & \multicolumn{1}{c|}{99} & 8.9 & 7.9 & 5.0 & 5.9 & 8.9 & - & 7.3 \\
 &  & Temporal Shuffle & \multicolumn{1}{c|}{93} & 74.2 & 5.0 & 8.9 & 5.0 & 5.9 & - & 19.8 \\
 &  & Diffpure\_DDPM & \multicolumn{1}{c|}{94} & \textbf{88.1} & \underline{80.2} & \textbf{79.2} & \underline{74.3} & \underline{52.5} & \textbf{77.2} & \underline{75.3} \\
 &  & Diffpure\_DDIM & \multicolumn{1}{c|}{93} & {91.1} & 62.4 & 68.3 & 61.3 & 19.8 & 39.6 & 57.1 \\
 &  & DDIM Inversion & \multicolumn{1}{c|}{97} & 33.7 & 6.9 & 17.8 & 5.9 & 7.9 & 13.9 & 14.4 \\
\multirow{-16}{*}{UCF-101} & \multirow{-8}{*}{SlowFast} & Ours & \multicolumn{1}{c|}{98} & \underline{75.2} & \textbf{90.1} & \underline{68.3} & \textbf{85.1} & \textbf{91.0} & \underline{63.4} & \textbf{78.9} \\ \midrule
 &  & None & \multicolumn{1}{c|}{100} & 5.0 & - & 5.0 & - & - & - & - \\
 &  & JPEG & \multicolumn{1}{c|}{97} & 7.0 & 6.0 & 6.0 & 7.0 & 6.0 & - & 6.4 \\
 &  & WD & \multicolumn{1}{c|}{99} & 7.0 & 5.0 & 6.0 & 6.0 & 4.0 & - & 5.6 \\
 &  & Temporal Shuffle & \multicolumn{1}{c|}{96} & 86.0 & 6.0 & 4.0 & 2.0 & 3.0 & - & 20.2 \\
 &  & Diffpure\_DDPM & \multicolumn{1}{c|}{98} & \textbf{93.0} & \underline{79.0} & \textbf{75.0} & \underline{65.0} & \underline{54.0} & \underline{66.0} & \underline{72.0} \\
 &  & Diffpure\_DDIM  & \multicolumn{1}{c|}{98} & 87.0 & 54.0 & 57.0 & 54.0 & {27.0} & 27.0 & 51.0 \\
 &  & DDIM Inversion & \multicolumn{1}{c|}{98} & 42.0 & 9.0 & 17.0 & 30.0 & 9.0 & 14.0 & 20.2 \\
 & \multirow{-8}{*}{NL} & Ours & \multicolumn{1}{c|}{91} & \underline{90.0} & \textbf{98.0} & \underline{61.0} & \textbf{73.0} & \textbf{98.0} & \textbf{81.0} & \textbf{83.5} \\ \cmidrule(l){2-11} 
 &  & None & \multicolumn{1}{c|}{100} & 4.0 & - & 5.0 & - & - & - & - \\
 &  & JPEG  & \multicolumn{1}{c|}{98} & 3.0 & 3.0 & 6.0 & 7.0 & 3.0 & - & 4.4 \\
 &  & WD & \multicolumn{1}{c|}{98} & 3.0 & 3.0 & 5.0 & 5.0 & 3.0 & - & 3.8 \\
 &  & Temporal Shuffle & \multicolumn{1}{c|}{92} & 80.0 & 4.0 & 5.0 & 3.0 & 3.0 & - & 19.0 \\
 &  & Diffpure\_DDPM & \multicolumn{1}{c|}{96} & \textbf{87.0} & \underline{71.0} & \textbf{72.0} & \underline{64.0} & \underline{44.0} & \textbf{71.0} & \underline{68.2} \\
 &  & Diffpure\_DDIM & \multicolumn{1}{c|}{95} & \underline{82.0} & 47.0 & 61.0 & 40.0 & {18.0} & \underline{26.0} & 45.7 \\
 &  & DDIM Inversion & \multicolumn{1}{c|}{96} & 26.0 & 3.0 & 16.0 & 12.0 & 3.0 & 14.0 & 12.3 \\
\multirow{-16}{*}{Kinetics-400} & \multirow{-8}{*}{SlowFast} & Ours &  \multicolumn{1}{c|}{93} & \underline{82.0} & \textbf{91.0} & \underline{66.0} & \textbf{82.0} & \textbf{93.0} & \textbf{71.0} & \textbf{80.8} \\ \bottomrule
\end{tabular}
\label{tab:main}
\end{center}
\end{table*}

\subsection{Performance Comparison}
In this section, we compare the defense performance of seven defense methods against eight adversarial attacks on two benchmark models and datasets. PGD and TT indicate gray-box attacks to evaluate Robust Acc* of defense methods. PGD+BPDA, TT+BPDA, EOT+BPDA and PGD+Diffattack, AutoAttack+BPDA indicate the adaptive attacks to evaluate Robust Acc of defense methods. We also select TT+MI as the black-box attack.
All attacks use the widely popular $l_\infty$ norm. For gray-box and adaptive attacks, $\epsilon_{adv}=4/255$, and for black-box attacks, $\epsilon_{adv}=16/255$.

\subsubsection{\textbf{PGD \& PGD+BPDA}} PGD is the most powerful first-order white-box attack method and is an effective tool to evaluate the performance of defense methods. As shown in Table~\ref{tab:main}, we find: 1) PGD shows strong attack performance on video models without defense methods. 2) JPEG and WD have almost no defense effect against PGD and PGD+BPDA. 3) Temporal Shuffle shows high Robust Acc* under PGD attack, but when facing the adaptive attack of PGD+BPDA, it shows weak defense performance. 4) Of the two methods in the Diffpure, Diffpure\_DDPM shows better performance due to the deeper gradient chain and stochasticity. 5) Compared to the second-best Diffpure\_DDPM, our method holds the Robust Acc* steady (a decrease of 3.8\% on average), and significantly enhances the Robust Acc (an increase of 19.1\% on average).

\subsubsection{\textbf{TT \& TT+BPDA}}
TT is specifically crafted for video recognition models. It disrupts the temporal information of videos by combining gradients from videos subjected to various shuffles. When left undefended, TT performs similarly to PGD in attacks. However, its attack efficacy against defense methods is notably enhanced. JPEG and WD remain the weak defense performance. Temporal Shuffle enhances defense by introducing randomness to input, yet TT's adaptability to different temporal patterns results in significant degradation compared to PGD. For instance, with NL/UCF101, Robust Acc* of Temporal Shuffle decrease by 71.2\% under TT. Our method outperforms Diffpure\_DDIM in both Robust Acc* and Robust Acc. In comparison to Diffpure\_DDPM, our Robust Acc* is lower but Robust Acc is higher by 13.6\% on average. It's worth noting that our method's defense performance against TT attacks shows a reduction compared to PGD. This is likely due to the distortion introduced by TT in the process of optical flow estimation, which can cause deviations during spatial-temporal optimization. In the next section, we will further explore the impact of attacks against optical flow on our defense methods.

\subsubsection{\textbf{EOT+BPDA}} 
EOT is explicitly tailored for defense mechanisms incorporating randomness. Consequently, there exists a notable decline in performance for methods such as Temporal Shuffle and the Diffpure series. Our method does not incorporate randomness throughout its entirety. VideoPure employs temporal DDIM inversion for the diffusion process and DDIM sampling for denoising. In essence, both the diffusing and denoising procedures rely on predicted noises by the denoising model. As a result, EOT's effectiveness is diminished against our method.

\subsubsection{\textbf{PGD+Diffattack}}
Diffattack is specifically devised to address challenges associated with inaccurate gradient estimation and memory overflow encountered in Diffpure's complex computational graphs, thereby achieving enhanced adaptive attack capabilities. Diffattack introduces a segment-wise forward-backward algorithm to enable memory-efficient gradient backpropagation. Consequently, when compared to PGD+BPDA, PGD+Diffattack demonstrates superior attack efficacy against most defense strategies owing to more precise gradient estimation. Furthermore, in contrast with various defense methods, our method maintains superior defense performance, surpassing Diffpure\_DDPM and Diffpure\_DDIM by margins of 6\%, and 47\% respectively.

\subsubsection{\textbf{AutoAttack}} We also verify the performance of different defense methods against stronger adaptive attacks, AutoAttack. Considering the large time cost (e.g. 10.3 hours for APGD-CE on one video), we randomly select 30 videos on UCF101 for testing. As shown in Table~\ref{tab:autoattack}, our method still maintains a significant defense performance under stronger adaptive attacks.

\begin{table}[!t]
\centering
\tabcolsep=1cm
\caption{Defense performance against AutoAttack}
\begin{tabular}{c|cccc@{}}
\toprule
Defense Method &  Robust Acc \\ \midrule
Temporal Shuffle  & 3.3 \\
Diffpure\_DDPM & 26.7 \\
Ours & \textbf{73.3} \\ \bottomrule
\end{tabular}
\label{tab:autoattack}
\end{table}

\subsubsection{\textbf{Black-box Attack}}
To comprehensively evaluate the performance of our method, we also investigate its defense performance against black-box attacks, where the attacker can not get  knowledge of the video recognition model and defense method. Specifically, we employ TT+MI~\cite{tt,mi} as black-box attack and conduct two experiments in two datasets, respectively: adversarial examples generated by NL to attack SlowFast and adversarial examples generated by SlowFast to attack NL, thereby simulating a black-box attack scenario. We adhere to the default black-box attack parameters with $\epsilon_{adv} = 16/255$ and attack step is 10. The average defense performance are presented in Table~\ref{tab:black}, revealing that our method maintains superior performance under black-box attack settings. Compared to the baseline DDIM inversion, our method demonstrates an 7.8\% improvement in black-box defense performance.

\begin{table}[!t]
\centering
\caption{Defense performance against black-box attacks.}
\begin{tabular}{@{}cccccc@{}}
\toprule
\multirow{2}{*}{None} & \multirow{2}{*}{\begin{tabular}[c]{@{}c@{}}JPEG\end{tabular}} & \multirow{2}{*}{\begin{tabular}[c]{@{}c@{}}Temporal\\ Shuffle\end{tabular}} & \multirow{2}{*}{\begin{tabular}[c]{@{}c@{}}Diffpure\\ DDPM\end{tabular}} & \multirow{2}{*}{\begin{tabular}[c]{@{}c@{}}DDIM\\ Inversion\end{tabular}} & \multirow{2}{*}{Ours} \\
 &  &  &  &  &  \\ \midrule
32.3 & 34.5 & 33.3 & 48.7 & 43.0 & \textbf{50.8} \\ \bottomrule
\end{tabular}
\label{tab:black}
\end{table}

\subsubsection{\textbf{Overall}} In summary, JPEG and WD exhibit minimal defense effectiveness across all models and datasets. Temporal Shuffle experiences notable performance decline when confronted with attacks targeting randomness or temporal aspects, and has weak defense performance against adaptive attacks. Diffpure\_DDPM outperforms Diffpure\_DDIM consistently across various attack methods, yet both exhibit substantial performance degradation against EOT attacks designed for randomness. Our method demonstrates robust defense performance against all attack types. Table~\ref{tab:main} showcases our method's optimal average defense performance across each model and dataset.

\subsection{Ablation Studies}
In this section, we undertake a series of experiments to validate the functionalities of various modules and designs within our method. All experiments are performed utilizing PGD and PGD+BPDA attacks ($l_\infty$ norm), employing the NL model, and the UCF-101 dataset.
\subsubsection{\textbf{Key Modules}}
We validate the efficacy of temporal DDIM inversion, spatial-temporal optimization, and multi-step voting in enhancing defense performance, as shown in Table~\ref{tab:keymodules}. Firstly, comparing the first and the second line, temporal DDIM inversion effectively suppresses the reconstruction ability of DDIM inversion towards the video, preserving the original video structure and cleaning adversarial noise, thereby achieving high Robust Acc* and Standard Acc. Secondly, comparing the second and the third line, we observe multi-step voting improves Robust Acc because of the existence of gradient estimation bias in adaptive attacks, wherein the model-estimated gradient may become trapped in the attack's local optimum for one of the candidate solutions ($z_0^t$) in the voting list. Regarding the roles of spatial-temporal optimization, when comparing the second and fourth lines, it primarily enhances resistance against adaptive attacks, resulting in a 59\% increase. However, this improvement comes at the cost of decreased Clean Acc and Robust Acc*. Finally, the combination of multi-step voting and spatial-temporal optimization mitigates the negative impact of spatial optimization on Clean Acc and Robust Acc* performance, thereby achieving robust defense performance.

\begin{table}[!t]
\centering
\caption{Ablation study on key modules.}
\label{tab:keymodules}
\scalebox{0.7}{
\begin{tabular}{@{}ccc|ccc@{}}
\toprule
\begin{tabular}[c]{@{}c@{}}Temporal \\ DDIM Inversion\end{tabular} & \begin{tabular}[c]{@{}c@{}}Spatial-temporal\\  Optimization\end{tabular} & Multi-step Voting & Standard Acc & Robust Acc* & Robust Acc \\ \midrule
 &  &  & \textbf{98.0} & 36.6 & 13.9 \\
\checkmark &  &  & 97.0 & 94.1 & 39.6 \\
\checkmark &  & \checkmark & 97.0 & \textbf{95.1} & 69.3 \\
\checkmark & \checkmark &  & 89.1 & 85.1 & 99.0 \\
\checkmark & \checkmark & \checkmark & 96.0 & 94.1 & \textbf{100} \\ \bottomrule
\end{tabular}}
\end{table}

\subsubsection{\textbf{Diffusion Model}}
To examine the influence of image diffusion models and video diffusion models on defense performance, we assess the efficacy of the latent diffusion model (Stable Diffusion V1.5~\cite{sd}) and pixel-level diffusion model (Guided Diffusion~\cite{guided}) in comparison to our adopted video diffusion model. As depicted in Table~\ref{tab:diffusion models}, whether employing Stable Diffusion or Guided Diffusion, the performance falls short of our adopted video diffusion model, particularly concerning Robust Acc. The weak defense performance of stable diffusion primarily results from significant disruption to the original image structure (notably, Diffpure\_DDPM achieves only 49.5\% Standard Acc). Although Guided Diffusion demonstrates respectable Standard Acc and Robust Acc*, it still exhibits inadequate defense capabilities against adaptive attacks. Modelscope, a video diffusion model tailored specifically for video input, excels in learning rich temporal information from videos. Consequently, following diffusion and denoising, it better preserves the original video structure (leading to improved Standard Acc). In summary, compared to the two image diffusion models, the video diffusion model consistently showcases superior defense performance.

 \begin{table}[!t]
\centering
\caption{Ablation study on diffusion models.}
\label{tab:diffusion models}
\scalebox{0.8}{
\begin{tabular}{@{}c|c|ccc@{}}
\toprule
Diffusion Model & Defense & Standard Acc & Robust Acc* & Robust Acc \\ \midrule
 & Diffpure\_DDPM & 49.5 & 30.7 & 22.8 \\
 \multirow{-2}{*}{Stable Diffusion} & Diffpure\_DDIM & 63.4 & 49.5 & 35.6 \\ \midrule 
 & Diffpure\_DDPM & 86.1 & 80.2 & 31.7 \\
\multirow{-2}{*}{Guided Diffusion} & Diffpure\_DDIM & 92.1 & 68.3 & 16.8 \\ \midrule
 & Diffpure\_DDPM & 94.1 & 88.2 & 72.3 \\
 & Diffpure\_DDIM & \textbf{96.0} & 85.1 & 41.6 \\
\multirow{-3}{*}{ModelScope} & Ours & \textbf{96.0} & \textbf{94.1} & \textbf{100} \\ \bottomrule
\end{tabular}}
\end{table}

\subsubsection{\textbf{Attacking Video Recognition and Optical Flow Models}}
To further confirm the efficacy of our method, we conduct attacks on both video recognition models and optical flow models to generate adversarial examples. Table~\ref{tab:attackboth} presents a performance comparison between our method and Diffpure. It is evident that the perturbation of optical flow does impact our method's performance. Nonetheless, when compared to Diffpure, our method consistently demonstrates superior defense performance.

\begin{table}[!t]
\center
\caption{Defense Performance when attacking both video recognition and optical flow models.}
\begin{tabular}{@{}c|cc@{}}
\toprule
Defense Method &  Robust Acc* & Robust Acc \\ \midrule
Diffpure\_DDIM &  86.1 & 67.3 \\
Diffpure\_DDPM &   88.1 & 79.2 \\
DDIM Inversion &  53.5 & 52.5 \\
Ours &  \textbf{89.1} & \textbf{97.0} \\ \bottomrule
\end{tabular}
\label{tab:attackboth}
\end{table}

\subsubsection{\textbf{Single-Step Prediction or Multi-step Voting}}
To further evaluate the efficacy of multi-step voting, we conduct a comparative experiment as shown in Table~\ref{tab:singlestep}. In this experiment, we select one candidate solution $z_0^t$ from the voting list to serve as the purified result and compare it with the multi-step voting approach. Our findings reveal that different candidate solutions have varying impacts on the final performance, with optimal results typically achieved from intermediate steps. we suggest that $z_0^t$ at the midpoint strikes an effective balance between preserving the original input's inherent structure and effectively eliminating adversarial noise. Ultimately, multi-step voting consistently demonstrates superior performance. 

\begin{table}[!t]
\caption{Defense performance of single step prediction or multi-step voting}
\centering
\label{tab:singlestep}
\begin{tabular}{@{}c|ccc@{}}
\toprule
Selected $z_0^t$ & Standard Acc & Robust Acc* & Robust Acc \\ \midrule
6 & 95.0 & 91.1 & 100 \\
5 & 96.0 & 91.1 & 100 \\
4 & 96.0 & 91.1 & 100\\
3 & 96.0 & 92.1 & 100\\
2 & 92.1 & 89.1 & 100\\
1 & 90.1 & 87.1 & 100\\ 
0 & 89.1 & 85.1 & 99.0\\ \midrule
Multi-step Voting & 96.0 & 94.1 & 100\\ \bottomrule
\end{tabular}
\end{table}

\subsubsection{\textbf{$z_{t}^{\prime}$ replaces $z_{t}^*$}}
In Algorithm~\ref{alg:algorithm} (line 6), the optimized $z_t^{\prime}$ is not directly substituted for $z_t^*$ during denoising. In other words, our optimization process does not directly influence the denoising procedure. Therefore, Table~\ref{tab:zt} examines the performance impact when $z_t^*$ is replaced with $z_t^{\prime}$ for denoising. It is evident that such a substitution significantly disrupts the normal denoising process, resulting in greater distortion to the original structure (Standard Acc decrease 17\%). Consequently, we opt to preserve the regular denoising trajectory and solely optimize the latent requiring voting.

\begin{table}[!t]
\centering
\caption{Ablation study on $z_t^{\prime}$ when $z_{t}^{\prime}$ replaces $z_{t}^*$. Ours* indicates $z_{t}^{\prime}$ replaces $z_{t}^*$ when denoising.}
\begin{tabular}{@{}c|ccc@{}}
\toprule
Defense Method & Standard Acc & Robust Acc* & Robust Acc \\ \midrule
Ours* & 79.2 & 66.3 &  76.2 \\
Ours & \textbf{96.0} & \textbf{94.1} & \textbf{100} \\ \bottomrule
\end{tabular}
\label{tab:zt}
\end{table}

\begin{figure}[!t]
\centering
  \scalebox{0.4}{
  \includegraphics[width=0.9\textwidth]{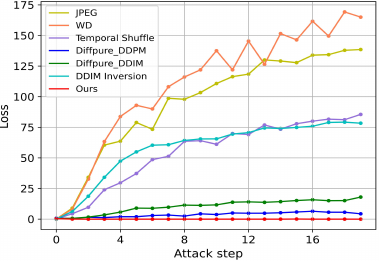}}
  \caption{Loss changes of different defense methods with attack steps under PGD+BPDA attack on NL model and UCF101 dataset.}
  \label{fig:loss_o}
\end{figure}

\begin{table}[!t]
\centering
\caption{Ablation study on loss of spatial-temporal optimization}
\begin{tabular}{@{}c|cc|ccc@{}}
\toprule
Attack& $L_{temp}$ & $L_{spa}$& Standard ACC & Robust Acc* & Robust Acc \\ \midrule
  \multirow{3}{*}{PGD}&  &  & \textbf{97.0} & \textbf{95.1} & 69.3 \\
 & & \checkmark & 96.0 & 94.1 & \textbf{100} \\
& \checkmark & \checkmark & 96.0 & 94.1 & \textbf{100} \\ \midrule
 \multirow{3}{*}{TT}&  &  & \textbf{97.0} & {54.5} & 49.5 \\
 & & \checkmark & 96.0 & 51.4 & \textbf{82.2} \\
& \checkmark & \checkmark & 96.0 & \textbf{55.4} & {80.2} \\ \bottomrule
\end{tabular}
\label{tab:loss}
\end{table}

\subsubsection{\textbf{Loss of spatial-temporal optimization}}
{To evaluate the effects of $L_{temp}$ and $L_{spa}$, we compare their defense performance against PGD and TT attacks, as summarized in Table~\ref{tab:loss}. The results indicate that $L_{spa}$ significantly enhances Robust Acc, i.e. the defense performance of adptive attacks. However, under the TT attack, which targets the temporal information in videos, using $L_{spa}$ alone results in a decrease in Robust Acc* (the defense performance of gray-box attacks) due to the lack of temporal information reconstruction. By incorporating $L_{temp}$, this temporal vulnerability is mitigated, with only a minor reduction in Robust Acc.}

\subsection{Effectiveness Analysis}
In this section, we analyze the effectiveness of our method in terms of defense performance when facing adaptive attacks. We choose PGD+BPDA as the adaptive attack method and NL as the video recognition model. During the adaptive attack process, the attacker estimates the overall gradient of the defense method and video recognition model at each iteration, thereby increasing the loss. On video recognition, this loss is cross-entropy loss. When the loss rises to a certain extent, the video recognition model will make incorrect judgments about the input. Therefore, a good adversarial defense method should be able to suppress this gradient rise. As shown in the Fig.~\ref{fig:loss_o}, we randomly select 20 videos from UCF101 for testing and calculate the mean change of the loss in each video during 20 iterations. Combining Table~\ref{tab:main}, we find the performance of the JPEG, WD, DDIM inversion, and Temporal Shuffle, which perform poorly, presents an apparent increase in loss, with just a few iterative steps making the loss skyrocket, thus rendering it incapable of effective defense. The relatively effective defense methods such as Diffpure\_DDPM slow down the rise of loss, but none of them successfully suppresses it. Our method successfully arrest the rise in loss, making the loss fluctuate within a small range, thus achieving a better adaptive defense performance.

\subsection{Time Cost}
{We verify the time cost comparison between our method and other diffusion-based methods on one NVIDIA GeForce 3090 GPU, as shown in Table~\ref{tab:time}. Compared to the second-best defense method, DiffPure\_DDPM, our method is approximately four times faster. Although our method is slightly slower than other DDIM-based methods, our method outperforms them significantly in terms of defense performance. In future work, we will further optimize our method to enhance its efficiency.}

\begin{figure}[!t]
\centering
  \scalebox{0.5}{
  \includegraphics[width=0.9\textwidth]{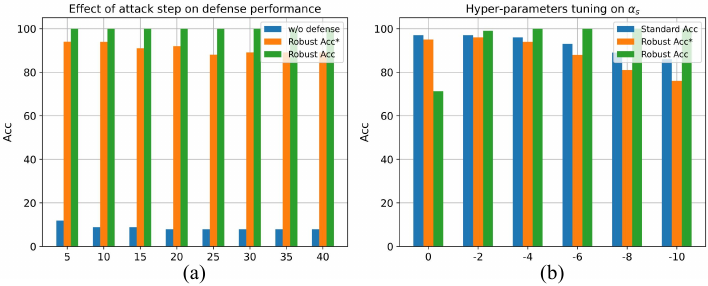}}
  \caption{(a) shows the effect of attack steps under PGD \& PGD+BPDA on NL model and UCF101 dataset. (b) shows hyper-parameters tuning on guidance scale.}
 \label{hyper}
\end{figure}

\begin{table}[!t]
\centering
\caption{Comparison of time cost and defense performance. Avg. Defense denotes the average defense performance across all models and datasets in Table~\ref{tab:main}.}
\label{tab:time}
\begin{tabular}{c|cccc@{}}
\toprule
Defense Method &  Time(s) & Avg. Defense(\%) \\ \midrule
Diffpure\_DDPM & 37.8 & 70.6 \\
Diffpure\_DDIM & \textbf{3.0} & 49.8 \\
DDIM Inversion & 5.2 & 17.0 \\
Ours & {8.5} & \textbf{82.3} \\ \bottomrule
\end{tabular}
\end{table}

\begin{figure}[!t]
\centering
  \scalebox{0.4}{
  \includegraphics[width=0.9\textwidth]{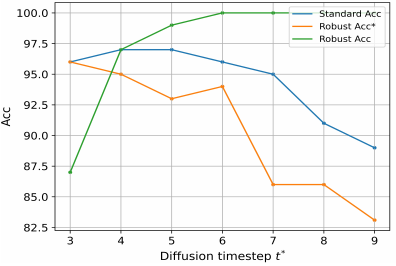}}
  \caption{Hyper-parameters tuning on diffusion timestep.}
  \label{fig:timestep}
\end{figure}

\subsection{Visualizations}
\begin{figure}[!t]
\centering
  \scalebox{0.51}{
  \includegraphics[width=0.9\textwidth]{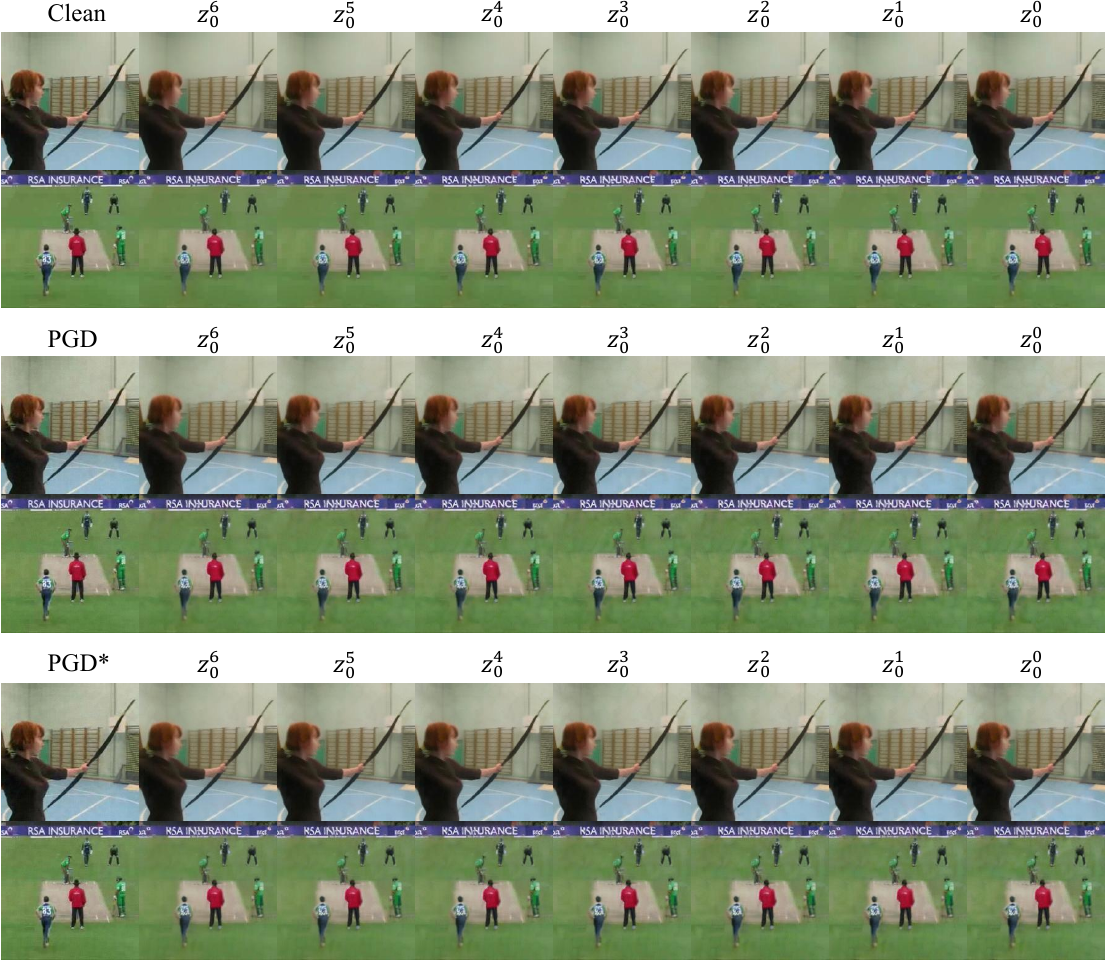}}
  \caption{
Visualizations of decoded $z^t_{0}$. The first two rows show purification for Standard Acc, the middle two for Robust Acc*, and the last two for Robust Acc. PGD* denotes PGD+BPDA.}
  \label{fig:vis}
\end{figure}

In Figure~\ref{fig:vis}, we present visualizations of different $z^t_0$ decoded into the pixel space. Since the final denoising steps are used, each $z^t_0$ effectively preserves the structure of the original frame while removing adversarial perturbations.

\subsection{Diagnostic Experiments \label{sec:hyper}}
In this section, we discuss the impact of different attack $\epsilon_{adv}$, attack steps and norm constraints on the defense performance of our method. In addition, we also conduct hyper-parameters tuning on guidance scale $\alpha_s$ and diffusion timestep $t^*$ of our method.

\begin{table}[t]
\caption{Defense performance against different $\epsilon_{adv}$}
\centering
\scalebox{0.9}{
\begin{tabular}{@{}c|c|ccc@{}}
\toprule
Defense Method& $\epsilon_{adv}$ & w/o  Defense & Robust Acc* & Robust Acc \\ \midrule
& 4 & 8.9 & 88.2 & 72.3 \\
& 8 & 5.9 & 82.2 & 40.6 \\
\multirow{-3}{*}{Diffpure\_DDPM} &16 & 2.0 & 67.3 & 32.7\\ \midrule
& 4 & 8.9 & 94.1 & 100 \\
& 8 & 5.9 & 85.1 & 100 \\
\multirow{-3}{*}{Ours} &16 & 2.0 & 73.3 & 100\\ \bottomrule
\end{tabular}}
\label{tab:eps}
\end{table}

\begin{table}[t]
\centering
\caption{Defense performance against $l_2$ attack}
\begin{tabular}{@{}c|ccc@{}}
\toprule
Defense Method & w/o  Defense & Robust Acc* & Robust Acc \\ \midrule
Diffpure\_DDIM & 21.8 & 93.1 & 89.1 \\
Diffpure\_DDPM & 21.8 & 93.1 & 93.1 \\
DDIM Inversion & 21.8 & 73.3 & 53.5\\
Ours & 21.8 & \textbf{96.0} & \textbf{97.0} \\ \bottomrule
\end{tabular}
\label{tab:l2}
\end{table}

\subsubsection{\textbf{Attack $\epsilon_{adv}$ \& Norm Constraint $l_2$ \& Attack Step}}
Table~\ref{tab:eps} illustrates the influence of attack $\epsilon_{adv}$ on defense performance. A larger perturbation range increases the denoising difficulty, resulting in a slight decline in our method's Robust Acc*, although without impacting our Robust Acc. Table~\ref{tab:l2} presents the defense performance of our method compared to other methods under the $l_2$ norm constraint. It is evident that varying norm constraints do not undermine the superiority of our method. Fig.~\ref{hyper} (a) depicts the defense performance of our method under different attack steps. We observe slight fluctuations in Robust Acc* while Robust Acc remains unaffected. In summary, we evaluate our defense method from three attack perspectives, further substantiating its generalization capability.

\subsubsection{\textbf{Diffusion Timestep $t^*$}}
Fig.~\ref{fig:timestep} shows the impact of diffusion timestep $t^*$ on the defense performance of our method. We find that the Standard Acc overall shows a downward trend, and the Robust Acc* also demonstrates a fluctuating downward trend with the timestep. This is because larger timestep, while providing greater coverage for adversarial noise, will also cause more substantial destruction to the original structure of the video, leading to a performance decrease. Robust Acc with respect to timestep shows an increasing trend, as larger timestep bring more candidate solutions for voting, further increasing the difficulty for the attacker to estimate the gradient. Considering all three metrics, we choose $t^*=6$.

\subsubsection{\textbf{Guidance Scale $\alpha_s$}}
Fig.~\ref{hyper} (b) shows the impact of different guidance scale on performance. The greater the degree of spatial-temporal optimization, the better performance on Robust Acc, but it affects the performance of Standard Acc and Robust Acc*.Considering the overall defense performance, we select $\alpha_s=-4$.

\section{Conclusion \label{cap:con}}
In this paper, we introduce VideoPure, a novel diffusion-based adversarial purification framework for video recognition models. Our framework enhances adversarial defense through temporal DDIM inversion, spatial-temporal optimization, and multi-step voting strategies. Temporal DDIM inversion preserves the clean video structure of the input, while the spatial-temporal optimization and multi-step voting strategies disrupt attack optimization, significantly enhancing the adversarial robustness of video recognition models. We validate the defense performance of our method against state-of-the-art (SOTA) attack methods under various attack settings. Our method demonstrates superior defense performance against image attack methods, video attack methods, attacks designed for randomness, and those targeting diffusion models. Additionally, we conduct extensive experiments to verify the effectiveness of each module and the overall efficiency of our method. In conclusion, our method is an effective and flexible adversarial defense tool that easily integrates into various video recognition models.

\bibliography{new}
\bibliographystyle{IEEEtran}

\end{document}